\documentclass[10pt]{article} 
\usepackage[accepted]{tmlr}

\usepackage{amsmath,amsfonts,bm}

\def\eqref#1{equation~\ref{#1}}

\def\1{\bm{1}}

\DeclareMathAlphabet{\mathsfit}{\encodingdefault}{\sfdefault}{m}{sl}
\SetMathAlphabet{\mathsfit}{bold}{\encodingdefault}{\sfdefault}{bx}{n}

\usepackage{hyperref}
\hypersetup{hidelinks}
\usepackage{url}
\usepackage{booktabs}
\usepackage{amsmath}
\usepackage{amssymb}
\usepackage{graphicx}

\title{JuryProbe: An Empirical Consensus-Risk Diagnostic for Routing Reference-Free Factuality Judge Panels to Grounded Verification}

\author{\name Tianxin Zhou \email zhoutx0@gmail.com \\
    \addr Independent Researcher
      \AND
      \name Ruixi Lin \email ruix.lin@northeastern.edu \\
      \addr Independent Researcher}

\def\month{08}  
\def\year{2026}
\def\openreview{\url{https://openreview.net/forum?id=dgBczhxcZY}}

\begin{document}

\maketitle
\begin{abstract}
Model-based evaluation systems increasingly use panels of inexpensive LLM judges to make accept-or-escalate decisions. In factuality settings, accepting a claim because several reference-free judges agree can create a hidden risk: agreement may arise from shared false-negative blind spots rather than independent evidence. We introduce \textsc{JuryProbe}, an empirical consensus-risk diagnostic for reference-free factuality judge panels, paired with a calibration-based routing policy. JuryProbe estimates panel-level consensus risk from a labeled calibration probe using false-negative-only (FN-only) judge correlation and false-consensus lift; when a panel is flagged high-risk, reference-free majority accepts are routed to the same judge panel with trusted references. Using frozen FEVER-supported claim pools, fixed-seed construction, and audited Number and Entity corruption families, we show that reference-free panels exhibit substantial correlated false negatives (FN-only correlations of 0.402 and 0.368; false-consensus lifts of 3.13$\times$ and 18.13$\times$). Under a trusted-reference best-case diagnostic, unanimous false consensus drops to zero both on minimal-pair FEVER corruptions and on non-minimal-pair scientific evidence. In flagged settings, the routed policy is, by construction, equivalent to grounding every reference-free majority accept (verified in 34/34 flagged splits). Its improvement therefore comes from accept-conditioned grounding, while the diagnostic’s operational role is to determine whether to activate this policy. A fixed, pre-specified rule produces both labels out of sample: it flags synthetic, benchmark-authored, and scientific claim families in 8–10 of 10 splits each, while standing down on the negative control in 0 of 10 splits. There, standing down avoids grounding roughly 28\% of deployment claims at a 0.004 absolute increase in false-accept rate. Stress tests with BM25-retrieved references and distribution-shifted deployment streams show that false-accept reduction persists under weak retrieval, while true-accept coverage degrades and stale stand-down labels require periodic labeled recalibration. JuryProbe provides no formal risk guarantee, nor do we establish reliable stand-down on natural panels; the supported contribution is an empirical diagnostic of high-risk panel error dependence.
\end{abstract}

\section{Introduction}
\label{sec:introduction}
In many model-evaluation pipelines, the most consequential action is not producing a score, but deciding whether an output should be accepted without further verification. A cheap reference-free judge panel makes this decision attractive: several models can be queried quickly, and majority or unanimous agreement appears to provide redundancy. Yet this deployment logic relies on an assumption that is rarely tested: agreement is informative only when judge errors are sufficiently independent. If the judges share the same false-negative blind spots, the panel may accept a corrupted claim precisely when it appears most confident. In this setting, agreement is not merely noisy evidence; it can become the event that triggers unsafe acceptance. 

This paper studies that failure mode as a measurement-and-routing problem. We introduce \textsc{JuryProbe}, an empirical consensus-risk diagnostic with calibration-based routing for reference-free factuality judge panels. Prior work has established LLM-as-a-judge evaluation as a practical paradigm \citep{zheng2023judging,wang2024pandalm,zhu2025judgelm}, and recent work has explored replacing single judges with panels of smaller models \citep{verga2024replacingjudgesjuriesevaluating}. JuryProbe asks a narrower deployment question: when should agreement from a cheap reference-free factuality panel be trusted, and when should it be routed to grounded verification?

JuryProbe estimates consensus risk from a calibration probe using two complementary panel-level statistics. FN-only correlation measures dependence: whether judges fail on the same corrupted claims. False-consensus lift measures consequence: whether unanimous false acceptance occurs more often than expected from the judges' marginal false-negative rates. Together, these statistics define a panel-level risk regime rather than a sample-level classifier. 

This framing differs from uncertainty or disagreement-based escalation \citep{geifman2017selective,pmlr-v97-geifman19a,jung2025trust}: the dangerous cases are precisely those in which the reference-free judges agree. Disagreement-based escalation cannot catch unanimous false acceptance by construction. JuryProbe therefore routes accept decisions when calibration shows that the panel is in a high-risk consensus regime, not merely when the judges disagree on a particular item. The diagnostic is empirical: it does not certify that an unflagged panel is safe, and we evaluate both rule outcomes explicitly. 

We evaluate JuryProbe on audited Number and Entity corruptions from frozen FEVER pools and six additional claim families. Reference-free panels exhibit substantial consensus risk, with no unanimous false consensus observed under trusted references. In held-out evaluation, gains come from accept-conditioned grounding: in flagged splits, the routed policy is by construction identical to grounding every reference-free majority accept, while the diagnostic only determines whether to activate it. On a negative control, the rule stands down in all splits at a small false-accept cost.

\textbf{This paper makes three contributions.}
\begin{itemize}
    \item We define \emph{consensus risk} for reference-free factuality judge panels and show that FN-only correlation and false-consensus lift are complementary panel-level signals that capture dependence and its deployment consequence, a failure mode that is invisible by construction to disagreement-based escalation.
    \item We provide a trusted-reference grounding diagnostic: holding the judge panel fixed, the unanimous false consensus observed under reference-free judging is not observed when the same judges are given trusted references, both for minimal-pair FEVER corruptions and for non-minimal-pair scientific evidence. This diagnostic represents a best-case assessment under trusted references rather than a causal isolation of why grounding improves reliability.
    \item We evaluate JuryProbe-Routed with an explicit policy-identity analysis: in every flagged split, it coincides with the no-estimator Ground-All-Accepts baseline (verified in 34/34 flagged splits), so the diagnostic's operational role is to determine whether accept-conditioned grounding is activated. A fixed, pre-specified rule produces both labels across eight claim families, while stand-down on a negative control avoids approximately 28\% of reference acquisitions.
\end{itemize}

Together, these results suggest that reliability in factuality judging can be improved by risk-aware grounding rather than by relying on reference-free agreement or simply adding more reference-free judges.

\section{Related Work}
\label{sec:related-work}

\subsection{LLM-as-a-Judge and Judge Panel}
\label{sec:related-work-llm-judge}

LLM-as-a-judge systems have become a practical alternative to human evaluation for open-ended model outputs. \citet{zheng2023judging} introduce MT-Bench and Chatbot Arena as scalable evaluation setting based on LLM judgments, while \citet{wang2024pandalm} and \citet{zhu2025judgelm} study automatic and fine-tuned LLM evaluators. \citet{verga2024replacingjudgesjuriesevaluating} further motivate panels of smaller, diverse judges as an alternative to a single expensive judge. JuryProbe builds on this deployment setting, but studies a different question: not whether judge panels are useful on average, but when reference-free panel agreement should no longer be treated as reliable evidence because judge errors may be correlated on the same factual corruptions.

\subsection{Judge Correlation and Dependence}
\label{sec:related-work-dependence}

The value of a judge panel depends on whether additional judges provide additional independent evidence. \citet{kohli2026judgeseffectivevotescorrelated} shows that nominally large LLM judge panels can yield far fewer effective independent votes when judge errors are correlated.

Our contribution is complementary: whereas \citet{kohli2026judgeseffectivevotescorrelated} studies judge independence as a property of panel composition, JuryProbe narrows the failure mode to correlated false negatives on corrupted factual claims, measures the downstream consequence through false-consensus lift, and evaluates a routed policy that sends high-risk accept decisions to grounded verification.

\subsection{Factuality Verification and Grounding}
\label{sec:related-work-factuality}

Factuality evaluation has been studied through atomic fact decomposition, hallucination detection, sampling-based consistency, and retrieval-augmented verification \citep{min2023factscore,wei2024longform,manakul2023selfcheckgpt,li2023halueval}.

JuryProbe builds on the observation that grounding can improve factuality judgments, but it does not propose a new factuality verifier. Instead, grounding is used as an intervention and an escalation target. The question is not whether grounded verification is useful in general, but when a cheap reference-free panel should be routed to it.

\subsection{Selective Evaluation and Escalation}
\label{sec:related-work-escalation}
Selective prediction studies how models can abstain or defer on unreliable inputs, trading coverage for reliability \citep{geifman2017selective,pmlr-v97-geifman19a}. In LLM evaluation, \citet{jung2025trust} study escalation based on estimated agreement with human judgment. JuryProbe likewise treats evaluation as a decision problem, but differs in the routing signal: it routes based on measured consensus risk rather than uncertainty, confidence, or disagreement. This distinction matters because disagreement-based escalation cannot catch unanimous false acceptance by construction. These methods are complementary: \citet{jung2025trust} and \citet{geifman2017selective} give formal guarantees for judge--human agreement and single-model selective prediction, respectively, whereas JuryProbe studies panel error dependence without distribution-free guarantees. Its risk thresholds were fixed before held-out evaluation (Section~\ref{sec:consensus-risk}).

A compact comparison with related work is provided in Appendix~\ref{app:related-work-positioning}.

\section{JuryProbe Framework}
\label{sec:framework}

JuryProbe is an empirical consensus-risk diagnostic, paired with a routing policy, for reference-free factuality judging. Its goal is not to replace a judge panel with a new factuality verifier, but to decide when agreement from a reference-free panel should no longer be treated as sufficient evidence for accepting a claim.

\subsection{Problem Setup}
\label{sec:problem-setup}

We consider a factuality decision setting in which a claim must be either accepted as factual or rejected as non-factual. Each example consists of a claim $c_i$ and a ground-truth factuality label $y_i \in \{0,1\}$, where $y_i=1$ denotes a factual claim and $y_i=0$ denotes a corrupted claim. When grounded verification is used, a trusted reference $r_i$ is additionally provided. 

A judge receives a claim and returns a binary decision. For a panel of $m$ judges, let $z_{ij}^{\mathrm{RF}} \in \{0,1\}$ denote whether judge $j$ accepts item $i$ in the reference-free setting, where $1$ means accept and $0$ means reject. Throughout this work, an accept decision means that the judge treats the claim as factual. The panel majority decision is 

\begin{equation} A_i^{\mathrm{RF}} = \mathbb{I}\left[\sum_{j=1}^{m} z_{ij}^{\mathrm{RF}} \geq \left\lceil \frac{m}{2} \right\rceil \right].
\label{eq:rf-majority-acceptance}
\end{equation}

In our main experiments, $m=3$, so majority acceptance means that at least two judges accept the claim.

The critical failure mode is false acceptance of corrupted claims. For a corrupted item, we use the term false negative in the corruption-detection sense: the judge fails to detect the corruption and incorrectly accepts the claim as factual. Thus, for a corrupted item ($y_i=0$), judge $j$ makes a false negative when $z_{ij}^{\mathrm{RF}}=1$. A false-consensus event occurs when all judges accept the same corrupted claim:

\begin{equation}
C_i^{\mathrm{RF}} = \mathbb{I}\left[y_i=0 \ \wedge\  \sum_{j=1}^{m} z_{ij}^{\mathrm{RF}} = m \right].
\label{eq:rf-false-consensus}
\end{equation}

Disagreement-based escalation cannot detect this event: the panel appears maximally reliable while all judges make the same error. JuryProbe is motivated by the observation that it can occur at rates exceeding the independent-error expectation.

\subsection{Consensus Risk}
\label{sec:consensus-risk}

JuryProbe estimates consensus risk on a calibration probe before applying any routed policy to deployment examples. The calibration probe consists of corrupted claims from a fixed corruption family and the reference-free decisions of a judge panel on those claims. The resulting risk estimate is a panel-level statistic: it characterizes whether a particular judge panel tends to share false-negative failures on a class of factual corruptions. It is not a classifier that predicts whether an individual deployment item is safe. Let \(\mathcal{N}_{\mathrm{cal}} = \{\, i : y_i = 0 \,\}\) denote the corrupted items in the calibration probe. For each corrupted item $i\in\mathcal{N}_{\mathrm{cal}}$ and judge $j$, define the false-negative indicator $e_{ij} = \mathbb{I}\left[z_{ij}^{\mathrm{RF}}=1\right]$.

JuryProbe uses two complementary statistics. First, FN-only correlation measures whether judges fail on the same corrupted items. For each judge pair $(j,k)$, we compute the Pearson correlation between their false-negative vectors: \(\rho_{jk} = \mathrm{corr}\left(e_{\cdot j}, e_{\cdot k}\right).\) Because $e_{ij}$ is binary, this equals the phi coefficient. The panel-level FN-only correlation is the average pairwise correlation, 
\begin{equation}
    \rho_{\mathrm{FN}} =
\frac{2}{m(m-1)}\sum_{1 \leq j < k \leq m} \rho_{jk}.
\label{eq:fn-correlation}
\end{equation}
Positive FN-only correlation indicates that judges tend to miss the same corruptions rather than making independent errors.

Second, false-consensus lift measures the consequence of that dependence for unanimous false acceptance. Let $q_{\mathrm{obs}} = \frac{1}{|\mathcal{N}_{\mathrm{cal}}|}\sum_{i \in \mathcal{N}_{\mathrm{cal}}} \mathbb{I}\left[\sum_{j} e_{ij}=m\right]$ denote the observed false-consensus rate and $p_j$ each judge's marginal false-negative rate; under independent errors the expected rate is $q_{\mathrm{ind}} = \prod_{j=1}^{m} p_j$.
False-consensus lift is then 

\begin{equation}
L_{\mathrm{FC}}
=
\frac{q_{\mathrm{obs}}}
     {q_{\mathrm{ind}}} 
\label{eq:false-consensus-lift}
\end{equation}

Values above $1$ indicate excess unanimous false acceptance relative to independence; thus, lift captures the consequence of dependence rather than dependence itself. Together, the two statistics define a consensus-risk regime for a judge panel. Importantly, consensus risk is assessed at the panel level and remains fixed throughout deployment; it is not recomputed for individual claims.

To assess whether the observed rate exceeds marginal expectations test (3000 permutations) that preserves each judge's marginal false-negative rate while shuffling false-negative locations across items. The $p$-value is the fraction of shuffled panels whose unanimous false-consensus rate is at least as large as the observed rate, with one-count smoothing.

A panel is treated as high-risk when the calibration probe satisfies three conditions:
\[\rho_{\mathrm{FN}} > 0.15,\quad
L_{\mathrm{FC}} > 1.5,\quad
p < 0.05.\]

These thresholds define a risk regime, not an item-level decision rule. They were selected prior to deployment evaluation and are examined through a threshold-sensitivity analysis in Section~\ref{sec:robustness}. The risk regime is estimated only on calibration data and then carried forward to deployment evaluation. In the JuryProbe policy, the high-risk label means that reference-free agreement should no longer be treated as sufficient evidence for accepting claims without grounding. 

\textbf{Threshold provenance.} The three constants were selected from initial Number-family calibration diagnostics and frozen on 2026-06-09, before held-out multiseed evaluation. Thus, Number is the threshold-development setting, while Entity, Attribute, both controls, FEVER-Refutes, SciFact, and CREAK were evaluated out of sample under the unchanged rule. No held-out outcome informed threshold selection. The rule is empirical, not a guarantee: it cannot target a specified error rate, and the permutation test does not bound deployment error.

\subsection{Grounded Verification}
\label{sec:grounded-verification}

Grounded verification defines the reference-augmented protocol used for the trusted-reference grounding diagnostic and routed escalation. The same judge panel evaluates the same claim with access to a trusted reference; judge identities, binary output space, and majority aggregation are kept fixed.

We denote reference-free decisions by $z_{ij}^{\mathrm{RF}}$ and grounded decisions by $z_{ij}^{\mathrm{G}}$. In the reference-free setting, judge $j$ receives only the claim $c_i$ and returns $z_{ij}^{\mathrm{RF}} \in \{0,1\}$. In the grounded setting, the same judge receives $(c_i,r_i)$, where $r_i$ is the trusted reference, and returns $z_{ij}^{\mathrm{G}} \in \{0,1\}$. The ground-truth label $y_i$ is not provided to the grounded verifier.

Grounded majority acceptance $A_i^{\mathrm{G}}$ and grounded false consensus $C_i^{\mathrm{G}}$ are defined analogously to Eqs.~\ref{eq:rf-majority-acceptance} and~\ref{eq:rf-false-consensus}, replacing $z_{ij}^{\mathrm{RF}}$ with $z_{ij}^{\mathrm{G}}$.

Section~\ref{sec:grounding-diagnostic} compares paired reference-free and grounded judgments under a fixed judge panel to assess whether the false-consensus events observed without references persist when trusted references are supplied. Since the references are trusted by construction, we treat this comparison as a \emph{trusted-reference best-case diagnostic}, rather than evidence for a mechanism-level account of why grounding improves judgments.

\subsection{JuryProbe-Routed Policy}
\label{sec:routed-policy}

JuryProbe uses the consensus-risk estimate to decide whether reference-free accept decisions require grounded verification. Let $h \in \{0,1\}$ denote the panel-level risk label estimated on the calibration probe, where $h=1$ indicates that the panel is high-risk according to the thresholds in Section~\ref{sec:consensus-risk}. This risk label is fixed before deployment evaluation. It is not recomputed for each deployment item.

For a deployment claim $i$, the reference-free panel first produces the majority decision $A_i^{\mathrm{RF}}$ defined in Eq.~\ref{eq:rf-majority-acceptance}. The final JuryProbe decision is 

\begin{equation}
D_i^{\mathrm{JP}} =
\begin{cases}
A_i^{\mathrm{RF}}, & h=0, \\
A_i^{\mathrm{G}}, & h=1 \ \wedge\ A_i^{\mathrm{RF}}=1, \\
0, & h=1 \ \wedge\ A_i^{\mathrm{RF}}=0.
\end{cases}
\label{eq:juryprobe-policy}
\end{equation}

\begin{figure}[t]
    \centering
    \includegraphics[width=0.75\linewidth]{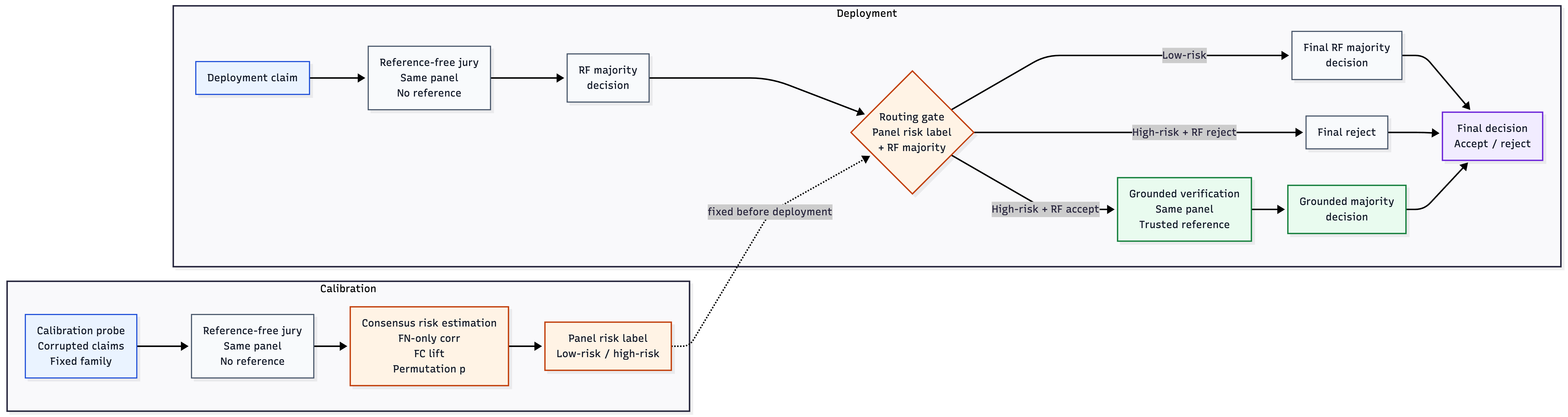}
    \caption{JuryProbe-routed policy. Consensus risk is estimated once on a calibration probe and fixed during deployment.}
    \label{fig:juryprobe-routed-policy}
\end{figure}

Figure~\ref{fig:juryprobe-routed-policy} summarizes the routed policy.

The asymmetry is deliberate: in factuality verification, false accepts are the safety-critical error, while rejects already block corrupted claims. JuryProbe therefore uses grounding only to guard against high-risk false accepts. Because routing can only overturn accepts, its false-accept rate cannot exceed the reference-free rate by construction. In contrast, full decision replacement offers no such protection and can increase false accepts (Section~\ref{sec:guardrail-evaluation}). Unlike disagreement-based escalation, JuryProbe can also route unanimous accepts, with verification driven by calibration rather than item-level disagreement.

\section{Experimental Setup}
\label{sec:experimental-setup}

We evaluate JuryProbe in a controlled factuality setting where clean claims are supported by trusted references and corrupted claims have known non-factual labels. This section describes the frozen corruption datasets, judge panels, metrics, and held-out protocol used to separate risk estimation from policy evaluation.

\subsection{Claim Pools and Corruption Families}
\label{sec:claim-pools}

We construct examples from FEVER-supported claims \citep{thorne-etal-2018-fever}. To avoid adaptive example selection, claim pools, corruption generation, and audits are frozen before judge evaluation.

The confirmatory evaluation uses two audited corruption families. Number corruptions modify numerical facts such as years, counts, or quantities, while Entity corruptions replace named entities with plausible alternatives. Each confirmatory family contains 300 clean and 300 corrupted examples, supporting equal-sized calibration and deployment splits. This evaluation budget was fixed before judge evaluation.

We additionally construct an Attribute family for replication analysis. Relation corruptions were explored but excluded before judge evaluation because audits revealed unstable fluency, syntax, and semantic-naturalness artifacts. 

Beyond FEVER-derived corruptions, we evaluate the fixed risk rule on five frozen families (Section~\ref{sec:cross-family}): two FEVER controls (Self-Contained Contradiction, Obvious-Number) and three benchmark sets (FEVER-Refutes, SciFact with 190 supported/190 contradicted claims \citep{wadden-etal-2020-fact}, and unfiltered CREAK \citep{onoe2021creak}). Construction details are in the appendix.

\subsection{Judge Panels}
\label{sec:judge-panels}

Our main judge panel consists of three relatively small LLM judges: Llama-3.1-8B-Instruct, Qwen-2.5-7B-Instruct, and Gemma-3-12B-IT. The same three judges are used for both reference-free judging and grounded verification. In the reference-free setting, judges receive only the claim; in the grounded setting, they receive the same claim with a trusted reference and determine whether the claim is fully consistent with that reference. Holding judge identities fixed keeps judge composition constant while the protocol changes by adding trusted references. All outputs are converted into binary accept/reject decisions. We additionally evaluate a pre-specified larger-capacity panel (Llama-3.3-70B-Instruct, Qwen-2.5-72B-Instruct, and Gemma-2-27B-IT) once on fixed SciFact claims, using a grouped five-fold protocol frozen before any judge calls (Section~\ref{sec:robustness}).

\subsection{Metrics}
\label{sec:metrics}

We report metrics for three purposes: consensus-risk estimation, the grounding diagnostic, and policy evaluation. 

For consensus-risk estimation, we report FN-only correlation ($\rho_{\mathrm{FN}}$), false-consensus lift ($L_{\mathrm{FC}}$), and the permutation-test $p$-value defined in Section~\ref{sec:consensus-risk}.

For the grounding diagnostic, we compare reference-free and grounded judging using FN-only correlation and unanimous false-consensus rate.

For policy evaluation, we report false accept rate (fraction of corrupted claims accepted), true accept rate (fraction of clean claims accepted), residual false-consensus rate (corrupted claims unanimously accepted reference-free and not blocked), and extra verifier calls. Wilson 95\% intervals and per-split ranges for key rates are given in the appendix.

\subsection{Held-out Evaluation Protocol}
\label{sec:heldout-protocol}

For policy evaluation, we separate risk estimation from policy measurement using held-out splits. For each corruption family and split seed, the 300 clean and 300 corrupted examples are partitioned into equal-sized calibration and deployment splits, each containing 150 clean claims and 150 corrupted claims. The calibration split is used only to estimate the panel-level risk label $h$ using the thresholds in Section~\ref{sec:consensus-risk}. The resulting risk label is fixed before deployment evaluation and is not recomputed on deployment examples.

We repeat this procedure over 10 split seeds and report mean and standard deviation across splits. This multi-seed protocol tests whether the high-risk label and policy outcomes are stable under different calibration/deployment partitions.

We compare JuryProbe-Routed against six baselines: Reference-Free Majority, Reference-Free Unanimity, Always Grounded, Ground-All-Accepts (grounds all RF majority accepts; equivalent to JuryProbe-Routed on high-risk splits), Disagreement-Routed, and Random-Routed (matched verifier budget). Missing grounded decisions are left missing rather than replaced with gold labels.

\section{Results}
\label{sec:results}

We structure the results into six parts: consensus-risk estimation, the trusted-reference grounding diagnostic, policy evaluation with identity and stand-down analyses, cross-family evaluation of the fixed rule, a retrieval stress test, and robustness analyses.

\subsection{Consensus Risk in Reference-Free Judging}
\label{sec:results-consensus-risk}

We first test whether reference-free judge panels exhibit consensus risk before any grounded intervention is applied. Table~\ref{tab:consensus-risk} reports the calibration statistics for the audited corruption families. FN-only correlation is computed over all corrupted claims. False-consensus rate, false-consensus lift, and permutation $p$-value are reported on a detectable corrupted subset. A corrupted claim is considered detectable if an external high-capability factuality detector (GPT-4o) correctly rejects the corruption.

\begin{table}[htbp]
\caption{Reference-free consensus-risk statistics. $N_{\mathrm{corr}}$ is the number of corrupted claims, $N_{\mathrm{det}}$ is the detectable corrupted subset used for difficulty-controlled false-consensus analysis, Corr. denotes FN-only correlation, and FC denotes false consensus.}
\centering
\small
\begin{tabular}{lrrrrrr}
\toprule
Family & $N_{\mathrm{corr}}$ & $N_{\mathrm{det}}$ & Corr. & FC Rate & FC Lift & $p$ \\
\midrule
Number & 300 & 214 & 0.402 & 0.159 & 3.13$\times$ & $<0.001$ \\
Entity & 300 & 286 & 0.368 & 0.031 & 18.13$\times$ & $<0.001$ \\
Attribute & 300 & 296 & 0.263 & 0.003 & 54.55$\times$ & 0.018 \\
\bottomrule
\end{tabular}
\label{tab:consensus-risk}
\end{table}

Both confirmatory families exhibit substantial consensus risk, with FN-only correlations of 0.402 and 0.368 and lifts of 3.13$\times$ and 18.13$\times$, respectively.

Attribute reproduces the signal ($\rho_{\mathrm{FN}}=0.263$, lift 54.55$\times$, $p=0.018$), but its low absolute false-consensus rate (0.003) limits its utility for paired grounding analysis.

Together, these results show that reference-free panel agreement can reflect correlated false-negative failures rather than independent confirmation, even when multiple judges agree. This motivates the next question: whether the false-consensus pattern remains when the same judges receive trusted references.

\subsection{Trusted-Reference Grounding Diagnostic}
\label{sec:grounding-diagnostic}

We next test whether the false-consensus pattern observed in reference-free judging remains when the same judges receive trusted references. This analysis uses the two confirmatory families, Number and Entity; Attribute is reported as replication evidence but is not used for this paired analysis, because its absolute false-consensus event rate is too low to provide a stable paired comparison.

Table~\ref{tab:grounding-collapse} compares reference-free and grounded judging on the same GPT-4o detectable corrupted claims. The reference-free statistics in this table are computed on that subset and therefore differ slightly from the all-corrupted FN-only correlation reported in Table~\ref{tab:consensus-risk}. The judge identities and aggregation rule are held fixed, while the protocol changes by adding trusted references.

\begin{table}[htbp]
\caption{Trusted-reference grounding diagnostic on the confirmatory families. RF denotes reference-free judging, Corr. denotes FN-only correlation, and FC denotes unanimous false consensus. The same judge panel evaluates the same corrupted claims with and without trusted references. Grounded FC lift is not reported because, in both families, the grounded false-consensus rate and the corresponding independence baseline are zero, making the lift ratio undefined.}
\centering
\small
\begin{tabular}{lrrrrr}
\toprule
Family & RF Corr. & Grounded Corr. & RF FC Rate & Grounded FC Rate & RF FC Lift \\
\midrule
Number & 0.386 & 0.000 & 0.159 & 0.000 & 3.13$\times$ \\
Entity & 0.393 & -0.003 & 0.031 & 0.000 & 18.13$\times$ \\
\bottomrule
\end{tabular}
\label{tab:grounding-collapse}
\end{table}

For both confirmatory families, the false-consensus pattern is not observed under grounding. Number falls from RF correlation 0.386 and false-consensus rate 0.159 to grounded correlation 0.000 and false-consensus rate 0.000. Entity shows the same pattern, with RF correlation 0.393 and false-consensus rate 0.031 falling to approximately zero under grounding.

These results provide a trusted-reference best-case diagnostic: with judges and aggregation fixed, the false-consensus pattern under reference-free judging is not observed with benchmark evidence. Because FEVER corruptions are minimal edits directly contradicted by their references, may partly reflect the simple consistency check.

Beyond minimal-pair edits, we evaluate 190 contradicted SciFact claims against the benchmark rationale and full published abstract (274 tokens on average). Neither reference was constructed from the claim. No three-judge false consensus is observed under either reference (0/190 vs.\ 10/190 reference-free), and per-judge false accepts drop substantially; mean pairwise FN correlation also falls but remains nonzero (0.247 with rationales; 0.108 with abstracts; Appendix~\ref{app:scifact-grounding}). Thus, grounding reduces but does not eliminate correlated errors, and these comparisons remain diagnostic rather than causal.

\subsection{Policy Evaluation: Baselines, Identity, and Stand-Down}
\label{sec:guardrail-evaluation}

We next evaluate whether consensus-risk routing improves deployment decisions. For each family, risk is estimated only on the calibration split and policy metrics are reported on held-out deployment examples over 10 split seeds. In both Number and Entity, JuryProbe detects a high-risk panel in all 10 splits.

\paragraph{Policy identity in flagged splits.} In every high-risk split, JuryProbe-Routed and Ground-All-Accepts are identical by construction: both ground exactly the reference-free majority accepts. We verify this on frozen caches: across all 34 flagged splits in which both policies were evaluated (Number 10, Entity 10, Attribute 8, boundary control 6), every reported field matches with zero discrepancies. Thus, improvements in flagged settings arise from accept-conditioned grounding, while the estimator's distinct operational role is determining whether to activate it. The policies differ only when the rule stands down, evaluated below on the negative control.

Table~\ref{tab:guardrail-baselines} compares JuryProbe-Routed with reference-free, grounded, no-estimator, disagreement-based, and budget-matched routing baselines. JuryProbe-Routed eliminates false accepts and residual false consensus in both confirmatory families ($0.427 \rightarrow 0.000$ for Number; $0.119 \rightarrow 0.000$ for Entity) while preserving the reference-free true-accept rate by routing accepts only.

\begin{table}[htbp]
\caption{Held-out policy evaluation over 10 split seeds. Values are mean $\pm$ standard deviation across splits. Residual FC denotes residual false-consensus rate after policy decisions. Verifier Calls denotes the number of grounded judge calls used by the policy. Always Grounded uses actual grounded verifier outputs from the same judge panel, not oracle labels. The Ground-All-Accepts rows match JuryProbe-Routed exactly because every split in both families is flagged high-risk, making the two policies identical by construction.}
\centering
\small
\setlength{\tabcolsep}{3pt}
\begin{tabular}{llcccc}
\toprule
Family & Policy & False Accept & True Accept & Residual FC & Verifier Calls \\
\midrule
Number & RF Majority & $0.427{\pm}0.029$ & $0.581{\pm}0.029$ & $0.201{\pm}0.022$ & $0{\pm}0$ \\
Number & RF Unanimity & $0.201{\pm}0.022$ & $0.273{\pm}0.027$ & $0.201{\pm}0.022$ & $0{\pm}0$ \\
Number & Disagreement-Routed & $0.201{\pm}0.022$ & $0.762{\pm}0.029$ & $0.201{\pm}0.022$ & $420{\pm}19$ \\
Number & Random-Routed & $0.212{\pm}0.013$ & $0.792{\pm}0.019$ & $0.100{\pm}0.009$ & $454{\pm}14$ \\
Number & Ground-All-Accepts & $0.000{\pm}0.000$ & $0.581{\pm}0.029$ & $0.000{\pm}0.000$ & $454{\pm}14$ \\
Number & \textbf{JuryProbe-Routed} & $\mathbf{0.000{\pm}0.000}$ & $0.581{\pm}0.029$ & $\mathbf{0.000{\pm}0.000}$ & $454{\pm}14$ \\
Number & Always Grounded & $0.000{\pm}0.000$ & $1.000{\pm}0.000$ & $0.000{\pm}0.000$ & $900{\pm}0$ \\
\midrule
Entity & RF Majority & $0.119{\pm}0.014$ & $0.639{\pm}0.027$ & $0.035{\pm}0.011$ & $0{\pm}0$ \\
Entity & RF Unanimity & $0.035{\pm}0.011$ & $0.394{\pm}0.027$ & $0.035{\pm}0.011$ & $0{\pm}0$ \\
Entity & Disagreement-Routed & $0.035{\pm}0.011$ & $0.841{\pm}0.015$ & $0.035{\pm}0.011$ & $306{\pm}10$ \\
Entity & Random-Routed & $0.074{\pm}0.008$ & $0.776{\pm}0.021$ & $0.022{\pm}0.007$ & $341{\pm}13$ \\
Entity & Ground-All-Accepts & $0.000{\pm}0.000$ & $0.639{\pm}0.027$ & $0.000{\pm}0.000$ & $341{\pm}13$ \\
Entity & \textbf{JuryProbe-Routed} & $\mathbf{0.000{\pm}0.000}$ & $0.639{\pm}0.027$ & $\mathbf{0.000{\pm}0.000}$ & $341{\pm}13$ \\
Entity & Always Grounded & $0.000{\pm}0.000$ & $1.000{\pm}0.000$ & $0.000{\pm}0.000$ & $900{\pm}0$ \\
\bottomrule
\end{tabular}
\label{tab:guardrail-baselines}
\end{table}

Disagreement-Routed leaves residual false consensus because unanimous false accepts provide no disagreement signal. Random-Routed fails to eliminate false accepts despite using the same verifier budget. Always Grounded eliminates false accepts but requires verification of every item. JuryProbe-Routed uses 49.6\% fewer verifier calls for Number and 62.1\% fewer for Entity, while accepting 58--64\% of clean claims versus 100\%. 

\paragraph{Stand-down on the negative control.} The fixed rule stands down in 10/10 Self-Contained Contradiction splits, issuing zero grounded calls. For an empirical no-estimator comparison, we grounded all reference-free-majority accepts appearing in at least one deployment split (165 unique items: 161 clean, 4 corrupted; 495 calls; zero parse failures) and ran Always Grounded on all 600 claims (1{,}800 verdicts; zero parse failures). Table~\ref{tab:standdown-control} reports the results.

\begin{table}[t]
\caption{Stand-down comparison on the Self-Contained Contradiction negative control (10 splits; 300-item deployment halves). All values are executed grounded-panel results, not oracle labels. References / Split counts unique claims requiring a trusted reference.}
\centering
\small
\begin{tabular}{lccc}
\toprule
Policy & False Accept & True Accept & References / Split \\
\midrule
RF Majority & $0.013{\pm}0.006$ & $0.552{\pm}0.023$ & $0$ \\
\textbf{JuryProbe-Routed} (stand-down) & $0.013{\pm}0.006$ & $0.552{\pm}0.023$ & $\mathbf{0}$ \\
Ground-All-Accepts & $0.009{\pm}0.005$ & $0.552{\pm}0.023$ & $84.7{\pm}4.0$ \\
Always Grounded & $0.138{\pm}0.025$ & $1.000{\pm}0.000$ & $300$ \\
\bottomrule
\end{tabular}
\label{tab:standdown-control}
\end{table}

Stand-down saves $84.7\pm4.0$ references per split (about 28\% of deployment claims) relative to Ground-All-Accepts, which reduces false accepts from 0.013 to 0.009. Always Grounded accepts 42/300 corrupted claims (false accept 0.138) while accepting all clean claims; this is the failure mode that one-sided accept protection avoids by construction (Section~\ref{sec:routed-policy}). An earlier oracle-label simulation suggesting 0.000 for Always Grounded on this control is documented and corrected in Appendix~\ref{app:utility}. This control demonstrates stand-down and its control-specific tradeoff, not natural-panel safety.

\paragraph{Boundary variance.} Attribute is flagged in 8/10 splits. In the two non-flagged splits (FN-correlation 0.083 and 0.119), JuryProbe retains reference-free decisions (false accept 0.047 and 0.040; residual unanimous false consensus 0.013), while Ground-All-Accepts would reduce false accepts to at most 0.013 and 0.007 at 127 and 117 references. This illustrates boundary variance near the 0.15 threshold.

\subsection{Cross-Family Evaluation of the Fixed Rule}
\label{sec:cross-family}

We apply the unchanged thresholds and splitwise estimator to eight claim families with 10 splits each, using disjoint calibration and deployment items (Table~\ref{tab:cross-family}). A split is flagged when all correlation, lift, and permutation criteria are met; mean lift is omitted when near-zero permutation nulls make it unstable, with per-split values in Appendix~\ref{app:cross-family-details}. Unlike Table~\ref{tab:consensus-risk}, which reports full-family statistics, this table is splitwise and calibration-side, so the aggregation scopes differ.

\begin{table}[t]
\caption{Cross-family evaluation of the fixed rule (frozen thresholds, no per-family tuning): flagged splits out of 10, with cross-split mean $\pm$ standard deviation of calibration-side FN-only correlation and lift. Entries marked -- are unstable under near-zero permutation nulls (Appendix~\ref{app:cross-family-details}).}
\centering
\small
\begin{tabular}{llrrr}
\toprule
Family & Type & Flagged & FN Corr & FC Lift \\
\midrule
Number & programmatic FEVER corruption & 10/10 & $0.385{\pm}0.028$ & $2.60{\pm}0.20$ \\
Entity & programmatic FEVER corruption & 10/10 & $0.348{\pm}0.062$ & $11.44{\pm}5.27$ \\
Attribute & programmatic FEVER corruption & 8/10 & $0.265{\pm}0.112$ & -- \\
SciFact & external scientific benchmark & 10/10 & $0.319{\pm}0.022$ & $5.57{\pm}0.70$ \\
FEVER-Refutes & benchmark-authored refutations & 8/10 & $0.306{\pm}0.073$ & $23.55{\pm}13.18$ \\
CREAK & benchmark-authored commonsense & 8/10 & $0.318{\pm}0.100$ & -- \\
Obvious-Number & boundary control & 6/10 & $0.330{\pm}0.167$ & -- \\
Self-Contained Contradiction & negative control & 0/10 & $0.082{\pm}0.054$ & 0.00 (9), undef.\ (1) \\
\bottomrule
\end{tabular}
\label{tab:cross-family}
\end{table}

The fixed rule yields both labels: signal families are flagged in 8--10/10 splits, the negative control in 0/10, and the boundary control in 6/10. A 95-clean/95-false SciFact sensitivity analysis remains flagged in 10/10 splits (FN correlation $0.305\pm0.043$; lift $5.11\pm0.78$). These results extend evaluation beyond the submitted FEVER corruptions but remain replication when calibration is drawn from the evaluated family; they do not establish generalization to error families absent from calibration, which we test under distribution shift in Section~\ref{sec:operating-regime}.

\subsection{Imperfect References: A BM25 Retrieval Stress Test}
\label{sec:retrieval-stress}

The zero false-accept rates in Table~\ref{tab:guardrail-baselines} use trusted benchmark FEVER evidence and are best-case results. We therefore stress-test routing on SciFact with a fixed, unsupervised, claim-only BM25 retriever over abstract sentences (corrupt-side any-evidence recall of 0.300 at $k{=}1$ and 0.437 at $k{=}3$). On the frozen subset (flagged in 10/10 splits, so routed rows equal Ground-All-Accepts), false accepts fall from 0.211 under reference-free majority to 0.016 (BM25@1) and 0.021 (BM25@3), versus 0.026 with benchmark rationales, while true-accept coverage falls from 0.595 to 0.295 and 0.326. The disjoint held-out analysis is consistent (BM25@3: false accept $0.020\pm0.023$, true accept $0.310\pm0.027$). Surviving unanimous false accepts fall from 10/190 to at most 2/190. Full results are in Appendix~\ref{app:bm25-details}. The reduction persists under weak retrieval, but does not establish BM25 sufficiency or robustness to incomplete, adversarial, or production references.

\subsection{Robustness and Replication}
\label{sec:robustness}

Four additional analyses (threshold sensitivity, grounded specificity, random-routing stability, judge-panel composition) confirm that the main findings are not driven by a single corruption family, estimator artifact, or a narrow threshold choice (Appendix~\ref{app:robustness-artifacts}). 

Two pre-specified evaluations probe natural settings. On fixed SciFact claims, both judge panels (Section~\ref{sec:judge-panels}) are flagged in 5/5 grouped folds; the larger panel yields 44/190 reference-free false accepts and 21/190 unanimous false consensus, while full-abstract Ground-All-Accepts reduces false accepts to 2/190 using 183 references (Appendix~\ref{app:scifact-panels}). CREAK is a boundary case: 8/10 splits flagged, with reference-free false accept $0.029\pm0.010$ and true accept $0.767\pm0.024$. We retain both outcomes without evaluating replacement panels; neither establishes reliable stand-down on natural data (Section~\ref{sec:limitations}).

\section{Discussion}
\label{sec:discussion}

This section discusses the operating regime, distribution shift, limitations, and future directions. 

\subsection{Operating Regime, Costs, and Distribution Shift}
\label{sec:operating-regime}

\paragraph{Costs.} LLM-call cost is negligible (${\sim}10^{-5}$ USD/claim); the main deployment cost is trusted-reference acquisition. Utility analysis (Appendix~\ref{app:utility}) shows that routing to grounding beats grounding every claim when reference cost is below 0.29--0.42$\times$ the value of a true accept, and beats reference-free operation when it is below 0.16--0.42$\times$ the cost of a false accept. On the negative control, stand-down is preferred unless reference cost is below $\sim$0.6\% of a false accept. Thus, JuryProbe is most useful when false accepts are costly and trusted references are available but too expensive to acquire universally, rather than as a uniform improvement over Always Grounded.

\paragraph{Distribution shift and recalibration.} A stand-down label can become stable under distribution shift. In a frozen stress test, an initial negative-control probe stands down in 0/10 splits; disjoint streams then replace increasing fractions of false claims with Number, Entity, or SciFact corruptions. At 0/10/25/50/100\% shift, a separate 95/95 recalibration audit flags 0/9/10/10/10 splits for Number, 0/3/6/6/10 for Entity, and 0/4/5/7/10 for SciFact (Appendix~\ref{app:recalibration-grid}). Without recalibration, reference-free false-accept rate reach 0.429 (Number), 0.111 (Entity), and 0.198 (SciFact) at 100\% shift. These results support periodic labeled recalibration, but not zero-shot drift detection or guarantees under arbitrary open-world drift.

\subsection{Limitations}
\label{sec:limitations}

\textbf{No formal guarantee.} JuryProbe provides no distribution-free guarantee. Its thresholds define an empirical protocol: users cannot specify a target error rate and obtain a rule guaranteed to meet it, and the permutation test provides no bound on deployment error. A split not flagged as high-risk is therefore not a zero-error certificate; it only indicates that the pre-specified calibration criteria for grounded routing were not satisfied.

\textbf{Natural stand-down and scope.} Stand-down is demonstrated as a specificity result on a constructed negative control. On natural data, CREAK represents a boundary case, while both SciFact panel evaluations are flagged in every fold. Thus, this work does not establish when agreement from a natural factuality panel can be trusted without grounding. Our conclusions are limited to short, self-contained binary factuality claims evaluated by small open-weight panels under the stated corruption and reference conditions. Consensus-risk estimates are family-dependent and may not transfer to unseen corruption types, long-form or free-form outputs, or multilingual claims. We exclude relation corruptions from judge evaluation because preliminary audits revealed unstable fluency, syntax, and semantic-naturalness artifacts.

The detectable-subset analysis uses GPT-4o as an external, analysis-time detector. Although it is not part of the JuryProbe panel, grounded verifier, or routing policy, its use is still a modeling choice. Appendix~\ref{app:alternative-detectability} therefore reports all-corrupted and grounded-detectable variants to assess whether the qualitative conclusions depend on this choice.

\textbf{Reference quality.} All grounded results outside Section~\ref{sec:retrieval-stress} should be interpreted as trusted-reference best-case diagnostics. The BM25 stress test considers one deliberately weak retriever on a single benchmark; behavior under degraded or adversarial reference sources beyond this setting remains untested. The negative control further demonstrates that grounded verification is effective for errors that can be resolved against the reference, but does not necessarily address errors arising from a claim's internal logic (Section~\ref{sec:guardrail-evaluation}).

\textbf{Diagnostic, not causal.} The paired comparisons hold the judge panel fixed while adding trusted references; they diagnose the reference-augmented protocol rather than isolate reference wording, prompt framing, or other protocol-level factors. Moreover, pairwise FN correlation persists under grounding on SciFact. The detectable-subset analysis reduces but cannot fully eliminate the influence of shared difficulty \citep{kohli2026judgeseffectivevotescorrelated}.

\subsection{Future Work}
\label{sec:future-work}

Long-form, free-form, and multilingual factuality settings are natural directions for future work, as is determining when natural judge panels can be reliably deemed safe to trust without grounding. Extending this framework beyond factuality, particularly for planning, forecasting and recommendation where trusted evidence may be incomplete, remains an open challenge. 

\section{Conclusion}
\label{sec:conclusion}

This paper introduced JuryProbe, an empirical diagnostic of consensus risk in reference-free factuality judge panels, together with a calibration-based routing policy. Rather than assuming that agreement implies reliability, JuryProbe assesses whether a panel exhibits correlated false-negative errors that can give rise to false consensus: a failure mode that disagreement-based escalation cannot detect by construction.

Our supported findings are as follows. Reference-free panels exhibit substantial false-negative dependence on audited Number and Entity corruptions, while the false-consensus pattern is not observed when the same judges are provided with trusted references, for both minimal-pair and non-minimal-pair evidence. In flagged settings, the routed policy is identical by construction to grounding every reference-free majority accept (verified in 34/34 flagged splits): the resulting improvement comes from accept-conditioned grounding, while the diagnostic determines whether that policy should be activated. A fixed, pre-specified rule yields both labels across eight claim families, standing down on a negative control in every split and avoiding roughly 28\% of reference acquisitions at a 0.004 increase in false-accept rate. False-accept reduction persists under a deliberately weak BM25 retriever, albeit with substantial coverage loss, and stale stand-down labels are corrected through periodic labeled recalibration under the evaluated distribution shifts.

These conclusions remain subject to important bounds: JuryProbe provides no formal risk guarantee, does not establish reliable stand-down for natural judge panels, and relies on trusted-reference best-case grounding outside the retrieval stress test. Within these limits, reliability in LLM factuality judging depends not only on individual judge accuracy or panel agreement, but also on the dependence structure of judge errors. Agreement alone should therefore not be treated as sufficient evidence for acceptance unless that error dependence is understood.

\section{Broader Impact Statement}
\label{sec:broader-impact}

Three cautions guide the use of this work. First, stand-down decisions are specific to the judge panel and data distribution and require recalibration after changes; stale labels cannot detect open-world shift (Section~\ref{sec:operating-regime}). Second, trusted references may be incomplete, incorrect, or manipulated; our retrieval stress test does not establish robustness to adversarial or production sources. Third, automated factuality judging should not replace expert review in high-stakes settings without domain-specific validation.

\bibliography{main}
\bibliographystyle{tmlr}

\appendix

\section{Related Work Positioning}
\label{app:related-work-positioning}

\begin{table}[htbp]
\caption{Positioning of JuryProbe relative to related work. Panel = LLM judge panel; Fact. = factuality-oriented evaluation; Grounded = grounded verification; Escal. = selective escalation; Consensus Risk = routing based on measured correlated-failure risk.}
\label{tab:related-work-positioning}
\centering
\small
\begin{tabular}{lccccc}
\toprule
Work & Panel & Fact. & Grounded & Escal. & Consensus Risk \\
\midrule
Kohli (2026)      & \checkmark & x & x & x & Partial \\
Trust or Escalate & \checkmark & x & Partial & \checkmark & x \\
PoLL              & \checkmark & x & x & x & x \\
LongFact          & x & \checkmark & x & x & x \\
SAFE              & x & \checkmark & \checkmark & x & x \\
FActScore         & x & \checkmark & \checkmark & x & x \\
HaluEval          & x & \checkmark & x & x & x \\
\textbf{JuryProbe}& \checkmark & \checkmark & \checkmark & \checkmark & \checkmark \\
\bottomrule
\end{tabular}
\end{table}

\section{Additional Robustness and Artifact Details}
\label{app:robustness-artifacts}

Table~\ref{tab:appendix-robustness} reports the robustness checks summarized in Section~\ref{sec:robustness}. These checks test whether the main conclusions depend on threshold choice, grounded-output specificity, random-routing variance, grounded evaluation integrity, or judge-panel composition. All checks are computed from frozen datasets and cached model outputs. Note that the grounded-specificity check is based on grounded verifier outputs: it shows that the estimator is not an always-on trigger, but does not test the reference-free stand-down branch. That branch is evaluated by the negative control in Section~\ref{sec:guardrail-evaluation}.

\begin{table}[htbp]
\caption{Additional robustness and artifact checks. Threshold sensitivity reports high-risk detections over correlation thresholds from 0.10 to 0.25 and lift thresholds from 1.25 to 2.00. Grounded specificity reports high-risk detections when the same risk estimator is applied to grounded verifier outputs.}
\centering
\small
\setlength{\tabcolsep}{4pt}
\begin{tabular}{p{0.27\linewidth}p{0.38\linewidth}p{0.27\linewidth}}
\toprule
Check & Result & Purpose \\
\midrule
Threshold sensitivity & Number: 10/10; Entity: 10/10 & Not threshold-fragile \\
Grounded specificity & Number: 0/10; Entity: 0/10 & Not always high-risk \\
Random-Routed stability & 100 trials per split & Stable budget-matched baseline \\
Grounded evaluation integrity & No oracle, no gold fallback, no missing imputation & Actual verifier outputs only \\
Strong judge slice & $\rho_{\mathrm{FN}}=0.252$, $L_{\mathrm{FC}}=2.17\times$, $p<0.001$ & Stronger RF judges do not remove risk \\
\bottomrule
\end{tabular}
\label{tab:appendix-robustness}
\end{table}

\section{Representative False-Consensus Cases}
\label{app:false-consensus-cases}

Table~\ref{tab:false-consensus-cases} shows representative false-consensus cases. In each case, all three reference-free judges accept the corrupted claim, even though the trusted reference contradicts the modified factual element. When the same judges receive the reference, grounded verification rejects the claim. These examples illustrate why disagreement-based routing is insufficient: the reference-free panel does not disagree on these corrupted claims; it unanimously accepts them.

\begin{table}[htbp]
\caption{Representative false-consensus cases. RF denotes reference-free judging, and G denotes grounded judging with the same judge panel and trusted reference.}
\centering
\small
\begin{tabular}{p{1.0cm} p{5.0cm} p{5.0cm} p{2.5cm}}
\toprule
Family & Corrupted claim & Reference & Outcome \\
\midrule
Number & Pacific Rim was released July 14, 2013. & Pacific Rim was released July 12, 2013. & RF: 3/3 accept; G: 0/3 accept \\
Number & Alex Rodriguez was suspended for 169 games. &
Alex Rodriguez was suspended for 211 games. &
RF: 3/3 accept; G: 0/3 accept \\
Entity & Steve Jobs's birth date is October 28, 1955. & Bill Gates's birth date is October 28, 1955. & RF: 3/3 accept; G: 0/3 accept \\
Entity & Elton John was named MusiCares' person of the year in 2013. &
Bruce Springsteen was named MusiCares' person of the year in 2013. & RF: 3/3 accept; G: 0/3 accept \\
\bottomrule
\end{tabular}
\label{tab:false-consensus-cases}
\end{table}

\section{Additional Experimental Details}
\label{app:experimental-details}

\subsection{Judge Prompts and Output Parsing}
\label{app:prompts-parsing}
All reference-free and grounded judgments are converted into binary decisions before evaluation. In the reference-free condition, each judge receives only the claim and is asked to decide whether the statement is factually correct. The prompt requires a one-word output, \texttt{true} if the statement is fully correct and \texttt{false} if it contains any factual error. In the grounded condition, each judge receives a trusted reference together with the claim and is asked whether the claim is fully consistent with the reference. The grounded prompt also requires a one-word \texttt{true}/\texttt{false} output.
Outputs are parsed using a strict string matcher for \texttt{true} or \texttt{false}. Responses that do not contain a valid binary verdict are marked as parse failures rather than interpreted manually. For the final grounded-verifier cache used in the policy evaluation, parse failures are retried with the same grounded prompt. The validated cache contains the expected number of grounded decisions for both confirmatory families: 1800 grounded judge outputs for Number and 1800 for Entity, corresponding to 600 examples times three judges. The final validated cache has zero remaining parse failures.
No oracle replacement, gold-label fallback, or missing-item imputation is used. Grounded decisions are produced by the same judge panel used in reference-free evaluation, with the only protocol change being the addition of the trusted reference to the judge input. The ground-truth label is never provided to the judge and is used only for evaluation.

\subsection{Corruption Audit Protocol}
\label{app:corruption-audit}
All corruption families are frozen before judge evaluation. Number and Entity are used as confirmatory families because author audits found them reliable enough for the main consensus-risk, grounding-diagnostic, and policy evaluations. Number corruptions modify numerical facts such as years, counts, or quantities, while Entity corruptions replace named entities with plausible alternatives.
Attribute is reported as an additional replication family because it satisfies the consensus-risk criteria but has a very low absolute false-consensus event rate, limiting its usefulness for the paired grounding diagnostic and policy evaluation. Relation corruptions were explored during construction but excluded from judge evaluation because audits revealed unstable fluency, syntax, and semantic naturalness artifacts. This exclusion was made before using Relation in any main judge-panel evaluation.

\subsection{Detectable Subset Analysis}
\label{app:detectable-subset}
False-consensus lift and the paired grounding diagnostic are reported on a detectable corrupted subset to reduce the influence of items that may be difficult or ambiguous even for a stronger external factuality detector. Detectability is estimated using GPT-4o as an analysis-time external detector. GPT-4o is not part of the JuryProbe judge panel, is not used as the grounded verifier, and is not used by the routed policy. A corrupted item is included in the detectable subset if GPT-4o correctly rejects the corrupted claim.
This analysis-only filter focuses the false-consensus analysis on corrupted claims whose factual error is externally detectable, while still evaluating whether the reference-free judge panel unanimously accepts them. The detectable subset contains 214, 286, and 296 corrupted examples for Number, Entity, and Attribute, respectively. FN-only correlation is reported over all corrupted claims, while false-consensus rate, false-consensus lift, and the permutation-test $p$-value are reported on the detectable corrupted subset. The paired grounding diagnostic uses the same detectable corrupted claims in paired reference-free and grounded conditions.

\subsection{Alternative Detectability Definitions}
\label{app:alternative-detectability}
The main analysis uses the GPT-4o detectable subset because this filter is independent of both the JuryProbe judge panel and the grounded verifier. To check that the conclusions do not depend on this particular filter, we also recompute the reference-free false-consensus statistics under two alternative subsets: all corrupted claims, and a grounded-detectable subset consisting of corrupted claims for which at least one grounded judge rejects the claim. The grounded-detectable variant is reported only as a robustness check, not as the primary subset for the grounding diagnostic, because it is defined using grounded verifier outputs.

\begin{table}[htbp]
\caption{Alternative detectability definitions. Across all detectability definitions, the qualitative conclusion remains unchanged: reference-free panel exhibits excess false consensus, while no grounded false consensus is observed.}
\label{tab:alternative-detectability}
\centering
\begin{tabular}{llrrrrr}
\toprule
Family & Subset & N & RF Corr. & RF FC Rate & RF FC Lift & Grounded FC \\
\midrule
Number & GPT-4o detectable & 214 & 0.386 & 0.159 & 3.13$\times$ & 0.000 \\
Number & All corrupted & 300 & 0.402 & 0.193 & 2.69$\times$ & 0.000 \\
Number & Grounded-detectable & 300 & 0.402 & 0.193 & 2.69$\times$ & 0.000 \\
Entity & GPT-4o detectable & 286 & 0.393 & 0.031 & 18.13$\times$ & 0.000 \\
Entity & All corrupted & 300 & 0.368 & 0.030 & 13.19$\times$ & 0.000 \\
Entity & Grounded-detectable & 300 & 0.368 & 0.030 & 13.19$\times$ & 0.000 \\
\bottomrule
\end{tabular}
\end{table}

\section{Grounding Diagnostic with Non-Minimal-Pair Scientific Evidence}
\label{app:scifact-grounding}

Table~\ref{tab:scifact-grounding} presents the full metrics for the SciFact grounding diagnostic summarized in Section~\ref{sec:grounding-diagnostic}: 190 contradicted claims evaluated under reference-free judging, with the benchmark-annotated rationale, and with the full published abstract (274 tokens on average, with no marked rationale span). These references were not constructed by editing the claims and encode semantic rather than single-token contradictions. For example, a false claim states that aPKCz causes tumour enhancement through glutamine metabolism, whereas the evidence shows that \emph{loss} of aPKCz enhances tumorigenesis and characterizes aPKCz as a metabolic tumor suppressor.

\begin{table}[t]
    \caption{SciFact grounding diagnostic (190 contradicted / 190 supported claims). Per-judge false-accept rates list the three judges in the order Llama / Qwen / Gemma.}
    \centering
    \small
    \begin{tabular}{lrrr}
    \toprule
    Metric & Reference-free & Benchmark rationale & Full abstract \\
    \midrule
    All-3 false consensus & 0.053 (10/190) & 0.000 (0/190) & 0.000 (0/190) \\
    FC lift & 5.56$\times$ & 0.00$\times$ & 0.00$\times$ \\
    Mean pairwise FN correlation & 0.312 & 0.247 & 0.108 \\
    Per-judge false accept & 0.23 / 0.10 / 0.43 & 0.07 / 0.01 / 0.06 & 0.06 / 0.02 / 0.06 \\
    Clean true accept (per-judge mean) & 0.598 & 0.737 & 0.791 \\
    Final parse failures & 1/1{,}140 & 0/1{,}140 & 0/1{,}140 \\
    \bottomrule
    \end{tabular}
    \label{tab:scifact-grounding}
\end{table}

\section{Cross-Family Evaluation: Split-Level Results and Full Statistics}
\label{app:cross-family-details}

Table~\ref{tab:cross-family-full} presents the complete cross-split statistics underlying Table~\ref{tab:cross-family}, including the lift values omitted from the latter and the corresponding permutation $p$-values. Lift is defined as a ratio relative to the independence null $q_{\mathrm{ind}}$; when the marginal false-negative rates are very low, $q_{\mathrm{ind}}$ approaches zero on the 95-item calibration halves, and a single false-consensus event can therefore yield an extreme ratio. As a result, cross-split mean lifts become unstable for Attribute, CREAK, and the boundary control; one boundary-control split even has an infinite lift. This instability does not affect the splitwise high-risk decision, which is determined by a per-split conjunction rather than by a cross-split mean. In the negative-control split containing a constant all-zero false-negative vector for one judge, pairwise FN correlations involving that judge are assigned 0.0 under the zero-variance convention used in the implementation, while lift is undefined because the corresponding independence baseline is zero.

\begin{table}[htbp]
\caption{Full cross-family splitwise statistics (mean $\pm$ standard deviation over 10 splits). Lift values marked unstable are dominated by near-zero independence nulls and are not comparable across families.}
\centering
\small
\setlength{\tabcolsep}{4pt}
\begin{tabular}{lrrrr}
\toprule
Family & Flagged & FN Corr & Lift & Permutation $p$ \\
\midrule
Number & 10/10 & $0.385{\pm}0.028$ & $2.60{\pm}0.20$ & $0.000{\pm}0.000$ \\
Entity & 10/10 & $0.348{\pm}0.062$ & $11.44{\pm}5.27$ & $0.011{\pm}0.016$ \\
Attribute & 8/10 & $0.265{\pm}0.112$ & $79.55{\pm}87.23$ (unstable) & $0.209{\pm}0.417$ \\
SciFact & 10/10 & $0.319{\pm}0.022$ & $5.57{\pm}0.70$ & $0.000{\pm}0.000$ \\
FEVER-Refutes & 8/10 & $0.306{\pm}0.073$ & $23.55{\pm}13.18$ & $0.016{\pm}0.031$ \\
CREAK & 8/10 & $0.318{\pm}0.100$ & $92.97{\pm}64.29$ (unstable) & $0.205{\pm}0.419$ \\
Obvious-Number & 6/10 & $0.330{\pm}0.167$ & $611.11{\pm}952.88$ ($+1$ inf; unstable) & $0.401{\pm}0.516$ \\
Self-Contained Contradiction & 0/10 & $0.082{\pm}0.054$ & $0.00$ (9 splits), undef.\ (1) & $1.000{\pm}0.000$ \\
\bottomrule
\end{tabular}
\label{tab:cross-family-full}
\end{table}

On CREAK, the reference-free deployment rates are false accept $0.029\pm0.010$ and true accept $0.767\pm0.024$. FEVER-Refutes has no trusted reference in our setup, so neither grounded policy is evaluable there; the Obvious-Number boundary control has no grounded cache, so its grounded-policy rates are analytic intervals rather than measured results and are not quoted as policy performance.

\section{Distribution-Shift Recalibration Grid}
\label{app:recalibration-grid}

Table~\ref{tab:recalibration-grid} presents the full stress-test grid summarized in Section~\ref{sec:operating-regime}. The initial calibration relies only on the negative control and stands down in 0/10 splits across all audit sizes. Each cell reports the number of splits (out of 10) flagged by the separate labeled recalibration audit, as a function of the target family's share of false claims in the shifted stream and the audit's labeled-item budget (25, 50, or 95 corrupted items). Stale RF FA denotes the reference-free false-accept rate on the shifted stream while the initial stand-down label is retained. The main-text counts report the 95-item audit column.

\begin{table}[htbp]
    \centering
    \caption{Recalibration flags (out of 10 splits) by target share and labeled-audit size, with the stale reference-free false-accept rate of the shifted stream.}
    \centering
    \small
    \begin{tabular}{llrrrr}
    \toprule
    Target family & Share & Audit 25 & Audit 50 & Audit 95 & Stale RF FA \\
    \midrule
    Number & 0\% & 0 & 0 & 0 & $0.013{\pm}0.008$ \\
    Number & 10\% & 2 & 7 & 9 & $0.054{\pm}0.032$ \\
    Number & 25\% & 6 & 8 & 10 & $0.116{\pm}0.038$ \\
    Number & 50\% & 7 & 10 & 10 & $0.217{\pm}0.044$ \\
    Number & 100\% & 8 & 10 & 10 & $0.429{\pm}0.051$ \\
    \midrule
    Entity & 0\% & 0 & 0 & 0 & $0.013{\pm}0.008$ \\
    Entity & 10\% & 0 & 1 & 3 & $0.021{\pm}0.015$ \\
    Entity & 25\% & 1 & 4 & 6 & $0.033{\pm}0.018$ \\
    Entity & 50\% & 4 & 4 & 6 & $0.063{\pm}0.023$ \\
    Entity & 100\% & 3 & 6 & 10 & $0.111{\pm}0.018$ \\
    \midrule
    SciFact & 0\% & 0 & 0 & 0 & $0.013{\pm}0.008$ \\
    SciFact & 10\% & 2 & 2 & 4 & $0.033{\pm}0.018$ \\
    SciFact & 25\% & 2 & 3 & 5 & $0.072{\pm}0.026$ \\
    SciFact & 50\% & 2 & 4 & 7 & $0.113{\pm}0.032$ \\
    SciFact & 100\% & 4 & 6 & 10 & $0.198{\pm}0.031$ \\
    \bottomrule
    \end{tabular}
    \label{tab:recalibration-grid}
\end{table}
This stress test evaluates listed benchmark shifts after a new labeled audit; it does not establish zero-shot safety under arbitrary open-world drift.

\section{SciFact Panels: Fold-level Results and Wilson Intervals}
\label{app:scifact-panels}

This appendix expands the SciFact panel evaluations of Section~\ref{sec:robustness}. Table~\ref{tab:scifact-wilson} reports pooled out-of-fold rates with Wilson 95\% intervals for both panels, and Table~\ref{tab:scifact-folds} reports fold-level calibration statistics. Ground-All-Accepts outcomes are actual grounded-panel decisions with the full published abstract as reference, never gold-label substitutions. A non-flagged fold would exercise the stand-down branch; none occurred.

\begin{table}[htbp]
    \caption{Pooled out-of-fold rates on 190 contradicted / 190 supported SciFact claims, with Wilson 95\% intervals.}
    \centering
    \small
    \begin{tabular}{llrrr}
    \toprule
    Panel & Metric & Events & Rate & Wilson 95\% \\
    \midrule
    Main & RF false accept & 40/190 & 0.211 & [0.159, 0.274] \\
    Main & RF clean accept & 113/190 & 0.595 & [0.524, 0.662] \\
    Main & RF all-3 false consensus & 10/190 & 0.053 & [0.029, 0.094] \\
    Main & Ground-All-Accepts false accept & 3/190 & 0.016 & [0.005, 0.045] \\
    Main & Ground-All-Accepts clean accept & 96/190 & 0.505 & [0.435, 0.576] \\
    \midrule
    Larger & RF false accept & 44/190 & 0.232 & [0.177, 0.297] \\
    Larger & RF clean accept & 139/190 & 0.732 & [0.664, 0.790] \\
    Larger & RF all-3 false consensus & 21/190 & 0.111 & [0.073, 0.163] \\
    Larger & Ground-All-Accepts false accept & 2/190 & 0.011 & [0.003, 0.038] \\
    Larger & Ground-All-Accepts clean accept & 121/190 & 0.637 & [0.566, 0.702] \\
    \bottomrule
    \end{tabular}
    \label{tab:scifact-wilson}
    \end{table}
    
    \begin{table}[htbp]
    \caption{Fold-level calibration statistics for both SciFact panels (grouped five-fold protocol; all folds flagged).}
    \centering
    \small
    \setlength{\tabcolsep}{4pt}
    \begin{tabular}{llrrrrrr}
    \toprule
    Panel & Fold & FN corr & All-3 count & Lift & $p$ & RF FA & GAA FA \\
    \midrule
    Main & 1 & 0.296 & 7 & 6.34 & 0.0003 & 13/38 & 0/38 \\
    Main & 2 & 0.357 & 10 & 5.91 & 0.0003 & 5/38 & 0/38 \\
    Main & 3 & 0.228 & 5 & 4.60 & 0.0023 & 12/38 & 1/38 \\
    Main & 4 & 0.372 & 10 & 5.98 & 0.0003 & 4/38 & 1/38 \\
    Main & 5 & 0.297 & 8 & 4.66 & 0.0003 & 6/38 & 1/38 \\
    \midrule
    Larger & 1 & 0.413 & 13 & 7.60 & 0.0003 & 14/38 & 1/38 \\
    Larger & 2 & 0.471 & 18 & 6.77 & 0.0003 & 7/38 & 1/38 \\
    Larger & 3 & 0.416 & 15 & 7.16 & 0.0003 & 12/38 & 0/38 \\
    Larger & 4 & 0.471 & 20 & 6.28 & 0.0003 & 6/38 & 0/38 \\
    Larger & 5 & 0.492 & 18 & 6.65 & 0.0003 & 5/38 & 0/38 \\
    \bottomrule
    \end{tabular}
    \label{tab:scifact-folds}
\end{table}

\section{BM25 Retrieval Stress-Test Details}
\label{app:bm25-details}

Table~\ref{tab:bm25-stress} reports the full reference-source comparison underlying Section~\ref{sec:retrieval-stress}, using the frozen subset (190 clean / 190 corrupted; flagged in 10/10 splits, so the routed rows also coincide with the Ground-All-Accepts rows). Descriptively, the BM25@3 verifier-majority false-accept rate is 0.012 on retrieval hits (83 false claims) and 0.047 on misses (107 false claims). Because hit status is not randomized, this stratification reflects an association rather than a mechanism test.

\begin{table}[htbp]
    \caption{SciFact reference-source stress test (full frozen subset). Surviving RF-unanimous FA counts corrupted claims unanimously accepted reference-free that also pass grounded majority (count in parentheses). Grounded Claims counts items routed to grounded verification.}
    \centering
    \small
    \begin{tabular}{lrrrr}
    \toprule
    Policy & False Accept & True Accept & Surviving RF-unanimous FA & Grounded Claims \\
    \midrule
    RF Majority & 0.211 & 0.595 & 0.053 (10) & 0 \\
    RF Unanimity & 0.053 & 0.368 & 0.053 (10) & 0 \\
    Routed + benchmark rationale & 0.026 & 0.479 & 0.005 (1) & 153 \\
    Routed + benchmark full abstract & 0.016 & 0.505 & 0.005 (1) & 153 \\
    Routed + BM25@1 & 0.016 & 0.295 & 0.005 (1) & 153 \\
    Routed + BM25@3 & 0.021 & 0.326 & 0.011 (2) & 153 \\
    \bottomrule
    \end{tabular}
    \label{tab:bm25-stress}
\end{table}

Table~\ref{tab:bm25-heldout} reports the within-split held-out check: the routed policy is recomputed on disjoint deployment halves (40 clean / 40 corrupted per split), while the risk flag is estimated using the calibration half only (flagged in 10/10 splits). Every full-subset figure falls within one split-level standard deviation of the held-out mean, and the policy has no fitted parameters beyond the binary flag.

\begin{table}[htbp]
    \caption{SciFact held-out deployment-half policy check versus full-subset rates.}
    \centering
    \small
    \begin{tabular}{lrrrr}
    \toprule
    Policy & Held-out FA & Full-subset FA & Held-out TA & Full-subset TA \\
    \midrule
    RF Majority & $0.193{\pm}0.041$ & 0.211 & $0.580{\pm}0.071$ & 0.595 \\
    Routed + rationale & $0.028{\pm}0.025$ & 0.026 & $0.490{\pm}0.076$ & 0.479 \\
    Routed + full abstract & $0.010{\pm}0.013$ & 0.016 & $0.502{\pm}0.080$ & 0.505 \\
    Routed + BM25@1 & $0.013{\pm}0.018$ & 0.016 & $0.278{\pm}0.025$ & 0.295 \\
    Routed + BM25@3 & $0.020{\pm}0.023$ & 0.021 & $0.310{\pm}0.027$ & 0.326 \\
    \bottomrule
    \end{tabular}
    \label{tab:bm25-heldout}
\end{table}

\section{When Is Consensus-Risk Routing Worth It? A Parameterized Utility Analysis}
\label{app:utility}

\subsection{Model}
For a deployment stream of $V$ claims with corrupted fraction $\pi$, define the per-claim utility of policy $P$ as
\[
U(P) = v_{\mathrm{TP}}\,\mathrm{TP}(P) - c_{\mathrm{FA}}\,\mathrm{FP}(P) - c_{\mathrm{ref}}\,\mathrm{Refs}(P) - c_{\mathrm{LLM}}(P) - c_{\mathrm{cal}}(P)/V,
\]
where $\mathrm{TP}(P) = (1-\pi)\,\mathrm{TA}_P$ and $\mathrm{FP}(P) = \pi\,\mathrm{FA}_P$ are obtained from the class-conditional accept rates measured in our experiments ($\pi = 0.5$ on the frozen splits; other class mixes simply reweight these rates under the assumption that the class-conditional rates remain stable). Here, $\mathrm{Refs}(P)$ denotes the measured number of trusted references acquired per claim; $c_{\mathrm{LLM}}(P)$ is the judge-call cost from the run logs ($3.78 \times 10^{-6}$ USD per call, or approximately $10^{-5}$ USD per claim for every policy); and $c_{\mathrm{cal}}$ is the one-time calibration cost, consisting of 300 labeled items plus $0.0034$ USD in reference-free calls. This calibration cost is incurred only by the routed policy and amortized over $V$. Rejections receive zero utility, so any loss in coverage corresponds to forgone $v_{\mathrm{TP}}$.

For an alternative policy ALT, let $\Delta\mathrm{TP} = \mathrm{TP}(\mathrm{ALT}) - \mathrm{TP}(\mathrm{JP})$, $\Delta\mathrm{FP} = \mathrm{FP}(\mathrm{JP}) - \mathrm{FP}(\mathrm{ALT})$, and $\Delta\mathrm{Refs} = \mathrm{Refs}(\mathrm{ALT}) - \mathrm{Refs}(\mathrm{JP}) > 0$. JuryProbe-Routed is preferred if and only if
\[
c_{\mathrm{ref}} > \left[\, v_{\mathrm{TP}}\,\Delta\mathrm{TP} + c_{\mathrm{FA}}\,\Delta\mathrm{FP} + \Delta c_{\mathrm{LLM}} + c_{\mathrm{cal}}/V \,\right] / \Delta\mathrm{Refs}.
\]
Thus, every decision boundary is linear in $(v_{\mathrm{TP}}, c_{\mathrm{FA}}, c_{\mathrm{ref}})$, with coefficients taken directly from the measured results; no parameters are fitted.

\subsection{Measured inputs (\texorpdfstring{$\pi=0.5$}{pi=0.5})}
In the flagged families, JuryProbe-Routed and Ground-All-Accepts are identical, as verified split by split, so a single entry represents both policies. For Number, RF majority has TA 0.581 / FA 0.427 / 0 references per claim; JuryProbe-Routed has TA 0.581 / FA 0.000 / 0.504 references per claim; and Always Grounded has TA 1.000 / FA 0.000 / 1.000. For Entity, RF majority has TA 0.639 / FA 0.119 / 0 references per claim; JuryProbe-Routed has TA 0.639 / FA 0.000 / 0.379; and Always Grounded has TA 1.000 / FA 0.000 / 1.000.

In the stand-down regime (negative control, not flagged in 10/10 splits), all entries are fully empirical. RF majority and JuryProbe have TA 0.552 / FA 0.013 / 0 references per claim; Ground-All-Accepts has TA 0.552 / FA 0.009 / 0.282; and Always Grounded has TA 1.000 / FA 0.138 / 1.000.

\subsection{Break-even boundaries}
In the flagged regime, JuryProbe-Routed is preferred to Always Grounded if and only if $c_{\mathrm{ref}} > 0.422\,v_{\mathrm{TP}}$ for Number (split range 0.382--0.466) or $c_{\mathrm{ref}} > 0.290\,v_{\mathrm{TP}}$ for Entity (0.265--0.316). It is preferred to RF majority if and only if $c_{\mathrm{ref}} < 0.423\,c_{\mathrm{FA}}$ for Number or $c_{\mathrm{ref}} < 0.157\,c_{\mathrm{FA}}$ for Entity. In both comparisons, the omitted terms are negligible ($\Delta c_{\mathrm{LLM}} \approx 10^{-5}$ USD and $c_{\mathrm{cal}}/(V\,\Delta\mathrm{Refs})$).

Together, these inequalities define a family-dependent operating band, $0.29\text{--}0.42\,v_{\mathrm{TP}} < c_{\mathrm{ref}} < 0.16\text{--}0.42\,c_{\mathrm{FA}}$. The band is non-empty only when the cost of a false accept is at least approximately 1$\times$ (Number) to 1.9$\times$ (Entity) the value of a true accept. Below this band, Always Grounded dominates; above it, reference-free operation dominates.

In the stand-down regime, JuryProbe, which uses zero references, is preferred to Ground-All-Accepts if and only if $c_{\mathrm{ref}} > 0.006\,c_{\mathrm{FA}}$, and to Always Grounded if and only if $c_{\mathrm{ref}} > 0.224\,v_{\mathrm{TP}} - 0.063\,c_{\mathrm{FA}}$. The negative coefficient on $c_{\mathrm{FA}}$ arises because empirical Always Grounded has a higher false-accept rate (0.138) than the reference-free panel (0.013) on this family; consequently, when $c_{\mathrm{FA}} \geq 3.6\,v_{\mathrm{TP}}$, JuryProbe is preferred regardless of reference cost.

Under the SciFact reference-quality variants, JuryProbe-Routed is preferred to Always Grounded if and only if $c_{\mathrm{ref}} > 0.216\,v_{\mathrm{TP}} - 0.009\,c_{\mathrm{FA}}$ for rationale references or $c_{\mathrm{ref}} > 0.184\,v_{\mathrm{TP}} - 0.009\,c_{\mathrm{FA}}$ for BM25@3.

\subsection{Correction of an earlier reported number}
An earlier version of this analysis reported the negative control's false-accept rate as 0.013 ``versus 0.000 for Always-Grounded.'' The 0.000 value came from an oracle-label policy simulation in which the Always Grounded decision was assigned the gold label rather than generated by the grounded judge panel, and was mistakenly presented as an empirical result. The executed Always Grounded condition on all 600 control claims (1{,}800 verdicts; zero parse failures) yields a false-accept rate of $0.138 \pm 0.025$ and a clean true-accept rate of $1.000 \pm 0.000$ over the ten held-out splits. On the full frozen set, 42/300 corrupted claims are accepted, while 300/300 clean claims are accepted.

The claim that freely available references make Always Grounded dominant therefore holds in the evaluated Number and Entity settings, but not universally: the control's appended contradictions are absent from, rather than contradicted by, the reference, and a consistency-based grounded check accepts many items that the reference-free panel rejects. All other negative-control results remain unchanged.

\subsection{Assumptions and scope}
The analysis assumes a linear, risk-neutral, and per-claim-additive utility, with no distribution-free guarantee. TA and FA are class-conditional rates measured on balanced splits; reweighting by $\pi$ therefore assumes that these rates remain stable across different class mixes. Boundary constants are point estimates based on the means across 10 splits, with split-level ranges reported separately. All rates are specific to the evaluated judge panel, claim families, and reference sources; recalibration is required following any change to the model, prompt, retrieval method, or data.

\section{Reproducibility}
\label{app:reproducibility}
\paragraph{Judge prompts.} The reference-free prompt is, verbatim:
\begin{quote}\small\ttfamily
You are checking a single statement for factual accuracy.\\
You are given ONLY the statement, with no reference answer.\\[2pt]
Statement: \{statement\}\\[2pt]
Is this statement factually correct? Consider any dates, numbers, and claims.\\
Respond with ONLY one word: true (if fully correct) or false (if it contains any factual error).\\[2pt]
Your answer:
\end{quote}
The grounded prompt is, verbatim:
\begin{quote}\small\ttfamily
You are verifying a statement against a trusted reference.\\[2pt]
Reference (trusted, correct): \{reference\}\\[2pt]
Statement to check: \{statement\}\\[2pt]
Is the Statement fully consistent with the Reference? Answer ONLY 'true' (consistent) or 'false' (contradicts the reference).\\[2pt]
Your answer:
\end{quote}

The main panel consists of Llama-3.1-8B-Instruct, Qwen-2.5-7B-Instruct, and Gemma-3-12B-IT, while the pre-specified larger panel consists of Llama-3.3-70B-Instruct, Qwen-2.5-72B-Instruct, and Gemma-2-27B-IT. All judges are queried through the OpenRouter API at temperature 0.0 with a 64-token output limit. Outputs are parsed using a strict \texttt{true}/\texttt{false} matcher; unparseable responses are recorded as parse failures and retried with the same prompt, with no imputation.

Held-out splits use seeds 1--10 with 150/150 calibration/deployment halves (SciFact uses the grouped five-fold protocol over 190/190 claims). The permutation seed is the same as the split seed, and permutation tests use 3{,}000 permutations with one-count smoothing. The risk thresholds ($\rho_{\mathrm{FN}} > 0.15$, $L_{\mathrm{FC}} > 1.5$, $p < 0.05$) were frozen on 2026-06-09. SciFact selection was frozen before judging: a label-blind atomic filter was applied to both classes, 190 of 208 available CONTRADICT candidates were sampled without replacement using seed 42, and supported claims were length-matched without model-assisted selection. The Self-Contained Contradiction control appends one of 40 distinct self-contained numeric contradictions (year ordering, magnitude, or arithmetic) to clean claims; FEVER-Refutes consists of claims labeled REFUTES by FEVER annotators; and the CREAK sample was drawn without content filtering.

All datasets, judge verdicts, and grounded verdicts are cached with dated freeze manifests. Grounded caches cover 600 items each for Number and Entity (1{,}800 verdicts per family), all 600 negative-control claims (1{,}800 verdicts), and every routed SciFact item under all four reference conditions. No oracle replacement, gold-label fallback, or missing-item imputation is used in any reported policy result.

\end{document}